\documentclass{notices}
 \usepackage{amsfonts,amssymb,amsmath,amscd}

 \usepackage{algorithm}
 \usepackage[noend]{algpseudocode}
 \usepackage{float}
 \usepackage{subcaption}
 \usepackage{smartdiagram}
 \makeatletter
 \def\BState{\State\hskip-\ALG@thistlm}
 \makeatother

 \usepackage{titlesec}
 \titleformat*{\section}{\large\bfseries}
 \titleformat*{\subsection}{\normalsize\bfseries}
 \titleformat*{\subsubsection}{\small\bfseries}

 \usepackage{graphicx}
 \graphicspath{ {./images/} }
 \usepackage{latexsym}
 \usepackage{upgreek}

 \usepackage{amsthm}
 \usepackage{enumitem}
 \usepackage{mathrsfs}
 \usepackage{color}
 
 \newcommand{\dkl}{D_{\mbox {\tiny{\rm KL}}}}

 \newcommand{\pmh}{p_{\mbox {\tiny{\rm MH}}}}
 \newcommand{\qpcn}{q_{\mbox {\tiny{\rm pCN}}}}
 \newcommand{\apcn}{a_{\mbox {\tiny{\rm pCN}}}}

 \newcommand{\M}{\mathcal{M}}

 \newcommand{\law}{\mathscr{L}}

 \newtheorem{theorem}{Theorem}[section]
 \newtheorem{Result}[theorem]{Result}

 \usepackage{amssymb}
 \usepackage{pifont}
 \newcommand{\cmark}{\ding{51}}%
 \newcommand{\xmark}{\ding{55}}%

 \title{
 	Mathematical Foundations of Graph-Based Bayesian Semi-Supervised Learning
 }

 \author{
 	N. Garc\'ia Trillos
 	\affil{
 		Assistant Professor, Department of Statistics, University of Wisconsin-Madison Madison, WI 53706, USA, garciatrillo@wisc.edu
 	}
 	\and 
 	D. Sanz-Alonso
 	\affil{
 		Assistant Professor, Department of Statistics, University of Chicago, Chicago, IL 60637, USA, sanzalonso@uchicago.edu
 	}
 	\and
 	R. Yang
 	\affil{
 		Graduate Student, Committee on Computational and Applied Mathematics, University of Chicago, Chicago, IL 60637, USA, yry@uchicago.edu
 	}
 }

\begin{document}
 	
 	\maketitle

 	In recent decades,  science and engineering have been  revolutionized by a momentous growth in the amount of available data. However, despite the unprecedented ease with which data are now collected and stored, \emph{labeling} data by supplementing each feature with an informative tag  remains to be challenging. Illustrative tasks where the labeling process requires expert knowledge or is tedious and time-consuming include labeling X-rays  with a diagnosis, protein sequences with a protein type, texts by their topic, tweets by their sentiment, or videos by their genre. In these and numerous other examples, only a few features may be manually labeled due to cost and time constraints. How can we best propagate label information from a small number of expensive labeled features to a vast number of unlabeled ones? This is the question addressed by semi-supervised learning (SSL).

 	This article overviews recent foundational developments on graph-based Bayesian SSL, a probabilistic framework for label propagation using similarities between features. SSL is an active research area and a thorough  review of the extant literature is beyond the scope of this article\footnote{This paper will appear in AMS Notices, which limits to $20$ the references per article. For this reason, we refer to  \cite{van2020survey,trillos2017consistency,sanz2020unlabeled} for further pointers to the literature.}.  Our focus will be on topics drawn from our own research that illustrate the wide range of mathematical tools and ideas that underlie the rigorous study of the  statistical accuracy and computational efficiency of graph-based Bayesian SSL. 
 	
 	\section{Semi-Supervised Learning (SSL)}
 	Let us start by formalizing the problem setting. Suppose we are given 
 	\begin{align}\label{eq:data}
 	\begin{split}
 	& \{(x_i,y_i) \}_{i=1}^n \quad \quad  \quad  \text{labeled data,}  \\
 	& \{x_i\}_{i=n+1}^N  \quad \quad \quad  \,\,\,\,\,  \text{unlabeled data},
 	\end{split}
 	\end{align}
 	where $(x_i,y_i)_{i=1}^n$ are independent draws from a random variable with joint law $\law(X,Y),$ and $\{x_i\}_{i=n+1}^N$ are independent draws from the marginal law $\law(X).$ We refer to the $x_i$'s as features and to the $y_i$'s as labels.\footnote{In machine learning, features and labels are often referred to as inputs and outputs, respectively.}
 	Due to the cost associated with labeling features, in SSL applications the number $n$ of labels is usually small relative to the number $N$ of features. The goal is then to propagate the few given labels to the collection of all given features. Precisely, we consider the problem of using all labeled and unlabeled data to
 	estimate the conditional mean function 
 	$$f_0(x):=\mathbb{E}[Y|X=x]$$ at the given features $\{x_i\}_{i=1}^N$. We call $f_0$ the \emph{labeling function}.
 	For ease of exposition, we will restrict our attention to regression and classification problems, where the labeled pairs are generated from: 
 	\begin{align}\label{eq:model}
 	\begin{cases}
 	Y = f_0(X) + \eta, \,\,\eta \sim  \mathcal{N}(0,\delta^2) &\text{regression},       \\
 	\mathbb{P}(Y = 1 |X) = f_0(X) &\text{classification},
 	\end{cases}
 	\end{align}
 	with $\delta$ known in the regression setting. 
 	Regression and classification are prototypical examples of SSL tasks with real-valued and discrete-valued labels, respectively. To streamline the presentation, we focus on \emph{binary} classification where there are only two distinct classes, labeled by $0$ and $1$. Several probabilistic models for binary classification are reviewed in \cite{bertozzi2018uncertainty}.  Extensions to non-Gaussian noise or multi-class classification can be treated in a similar fashion.

 	As its name suggests, SSL lies between \emph{supervised} and \emph{unsupervised} learning. In supervised learning, the goal is to use labeled data to learn the labeling function $f_0$, so that it may later be evaluated at new features. On the other hand, unsupervised learning is concerned with using unlabeled data to  extract important geometric information from the feature space, such as its cluster structure. SSL leverages \emph{both} labeled and unlabeled data in the learning procedure; all given features are used to recover the geometry of the feature space, and this geometric information is exploited to learn $f_0$.
 	\begin{table}[!htb]
 		\centering
 		\begin{tabular}{|c|c|c|c|c|c}
 			\hline
 			Task         & Labeled  & Unlabeled    \\
 			\hline
 			Supervised  &    \cmark  & \xmark  \\
 			\hline
 			Unsupervised & \xmark & \cmark \\
 			\hline
 			Semi-supervised & \cmark  & \cmark \\
 			\hline
 		\end{tabular}
 		\label{table:FLS}
 	\end{table}

 	Relying on unlabeled data to estimate the labeling function may seem counterintuitive at first. Indeed, the question of whether unlabeled data can enhance the learning performance in SSL has been widely debated, and different conclusions can be reached depending on the assumed relationship between the label generating mechanism and the marginal distribution of the features.  
 	We will tacitly adopt the \emph{smoothness assumption} ---often satisfied in applications--- that similar features should receive similar labels. In other words, we assume that the labeling function varies smoothly along the feature space.
 	Under such a  model assumption, one can intuitively expect that unlabeled data may boost the learning performance: uncovering the geometry of the feature space via the unlabeled data facilitates defining a  smoothness-promoting regularization procedure   for the recovery of the labeling function. The main idea of the graph-based Bayesian approach to SSL is to use a graph-theoretical construction to turn pairwise similarities between features into  a \emph{probabilistic} regularization procedure.

 	\section{Graph-Based Bayesian SSL}\label{ssec:BayesvsOpt}
 	We will take a Bayesian perspective to learn the restriction of $f_0$ to the features, denoted
 	$$f_N:=f_0|_{\{x_1,\ldots,x_N\}}.$$ We view $f_N$ as a vector in $\mathbb{R}^N,$ with coordinates $f_N(i) := f_0(x_i).$
 	In the Bayesian approach, inference is performed using a \emph{posterior distribution} over $f_N$, denoted $\mu_N.$ The posterior density $\mu_N(f_N)$ will be large for functions $f_N$ that are consistent with $(i)$ the given labeled data; and $(ii)$ our belief that similar features should receive similar labels. The posterior density is defined by combining a \emph{likelihood function} and a \emph{prior distribution} that encode, respectively, these two requirements:
 	\begin{equation}\label{eq:posterior}
 	\underbrace{\mu_N(f_N)}_{\text{posterior}}  \propto \underbrace{ L(f_N; y)}_{\text{likelihood}}  \underbrace{{\pi_N(f_N) }}_{ \text{prior}}.  
 	\end{equation}
 	 Here and elsewhere $y := \{y_1, \ldots, y_n \}$ is used as a shortcut for all given labels.
 	We next describe, in turn, the definition of likelihood and prior, followed by a discussion of how the posterior distribution is used to conduct inference in the Bayesian framework. 
 	
 	\paragraph{Likelihood Function}
 	The likelihood function encodes the degree of probabilistic agreement of a labeling function $f_N$ with the observed labels $y$, based on the model defined by \eqref{eq:data} and \eqref{eq:model}. The independence structure in \eqref{eq:data} implies that the likelihood factorizes as
 	\begin{align*}
 	L(f_N ; y ) &: = \mathbb{P} (y | f_N)  =  \prod_{i=1}^n \mathbb{P}(y_i|f_N),
 	\end{align*}
 	and the  Gaussian and binomial  distributional assumptions in \eqref{eq:model} give that
 	\begin{align} 
 	\mathbb{P}(y_i |f_N) &=
 	\begin{cases}
 	(2\pi\delta^2)^{-\frac12}e^{-\frac{|y_i-f_N(i)|^2}{2\delta^2}} & \text{regression,}\\
 	f_N(i)^{y_i} \left[1-f_N(i)\right]^{1-y_i}  & \text{classification.}  \\
 	\end{cases}
 	\end{align}
 	Note that the features are not involved in the definition of the likelihood function; in particular, the likelihood function does not depend on the unlabeled data.

 	\paragraph{Prior Distribution}
 	The prior encodes the belief that the labeling function should take similar values at similar features. We thus seek to design the prior so that its density $\pi_N(f_N)$ is large for functions $f_N$ that vary smoothly along the feature data. We will achieve this goal by defining the prior as a transformation of a Gaussian random vector $u_N$ as follows:
 	\begin{align*}
 	\pi_N &=
 	\begin{cases}
 	\law(u_N) \quad \quad & \text{regression,}\\
 	\law(\Phi(u_N)) \quad \quad & \text{classification.}  \\
 	\end{cases}
 	\end{align*}
 	Here $\Phi:\mathbb{R}\rightarrow (0,1)$ is a link function that ensures that in the classification setting the prior samples  take coordinate-wise values in $(0,1)$. Section \ref{sec:prior} will discuss how to use graph-based techniques to define the covariance structure of the latent vector $u_N$ so that $u_N(i)$ and $u_N(j)$ are highly correlated if the features $x_i$ and $x_j$ are similar. We term the approach ``graph-based'' because we will view each feature $x_i$ as a node of a graph, and use a matrix $W$ of pairwise similarities between features to define weighted edges. 
 	Then, the covariance of $u_N$ will be defined  using a \emph{graph-Laplacian} to penalize certain discrete derivatives of $u_N$.  In applications, the pairwise similarities often take the form  $W_{ij}= \mathcal{K} \bigl(\varphi(x_i),\varphi(x_j)\bigr)$, where $\varphi$ is a feature representation map that embeds the feature space in a suitable Euclidean space, and $\mathcal{K}$ is a kernel function such as the squared exponential $\mathcal{K}(s,t)=e^{-|s-t|^2}$, with $|\cdot |$ the Euclidean norm. 
 	
 	Note that labels are not used in the definition of the prior; instead, the prior is designed using pairwise similarities between all labeled and unlabeled features $\{x_i\}_{i=1}^N$.

 	\paragraph{Bayesian Inference}
 	Bayes's formula \eqref{eq:posterior} combines likelihood and prior to obtain the posterior distribution, used to perform Bayesian inference. Before moving forward, notice again that the prior is constructed solely in terms of the features, whereas only the labels enter the likelihood function. This insight will be important in later sections. 
 	
 	The posterior density $\mu_N(f_N)$ quantifies our degree of belief that $f_N$ is the true (restricted) labeling function that generated the given data. A natural point estimator for $f_N$ is hence the posterior mode, 
 	\begin{align}
 	\begin{split}
 	\widehat{f}_N &:= \underset{g\in\mathbb{R}^{N}}{\operatorname{arg\,max}} \, \mu_N(g) \\
 	&= \underset{g\in\mathbb{R}^{N}}{\operatorname{arg\,max}} \, \log L(g ; y) + \log \pi_N(g).
 	\end{split}
 	\end{align}
 	The right-hand side showcases that the posterior mode, also known as the \emph{maximum a posteriori} estimator, can be found by optimizing an objective function comprising a \emph{data misfit} and a \emph{regularization} term, defined by the log-likelihood function and the log-prior density, respectively. This observation reconciles the Bayesian approach with classical optimization methods that  ---without a probabilistic interpretation--- recover the labeling function by minimizing an objective that comprises data misfit and regularization terms. 
 	
 	Under the Bayesian framework, however, the posterior mean and the posterior median can also be used as meaningful point estimators that can be robust to outliers or model misspecification. Moreover, in addition to enabling point estimation, the posterior distribution also allows  one  to quantify the uncertainty in the reconstruction of the labeling function by computing Bayesian confidence intervals, correlations, or quantiles. All these quantities can be expressed as expectations with respect to the posterior distribution. As will be detailed later, sampling algorithms such as Markov chain Monte Carlo may be used to approximate these posterior expectations.

	\begin{figure}[!htb]
		\centering
		\includegraphics[width=1\linewidth]{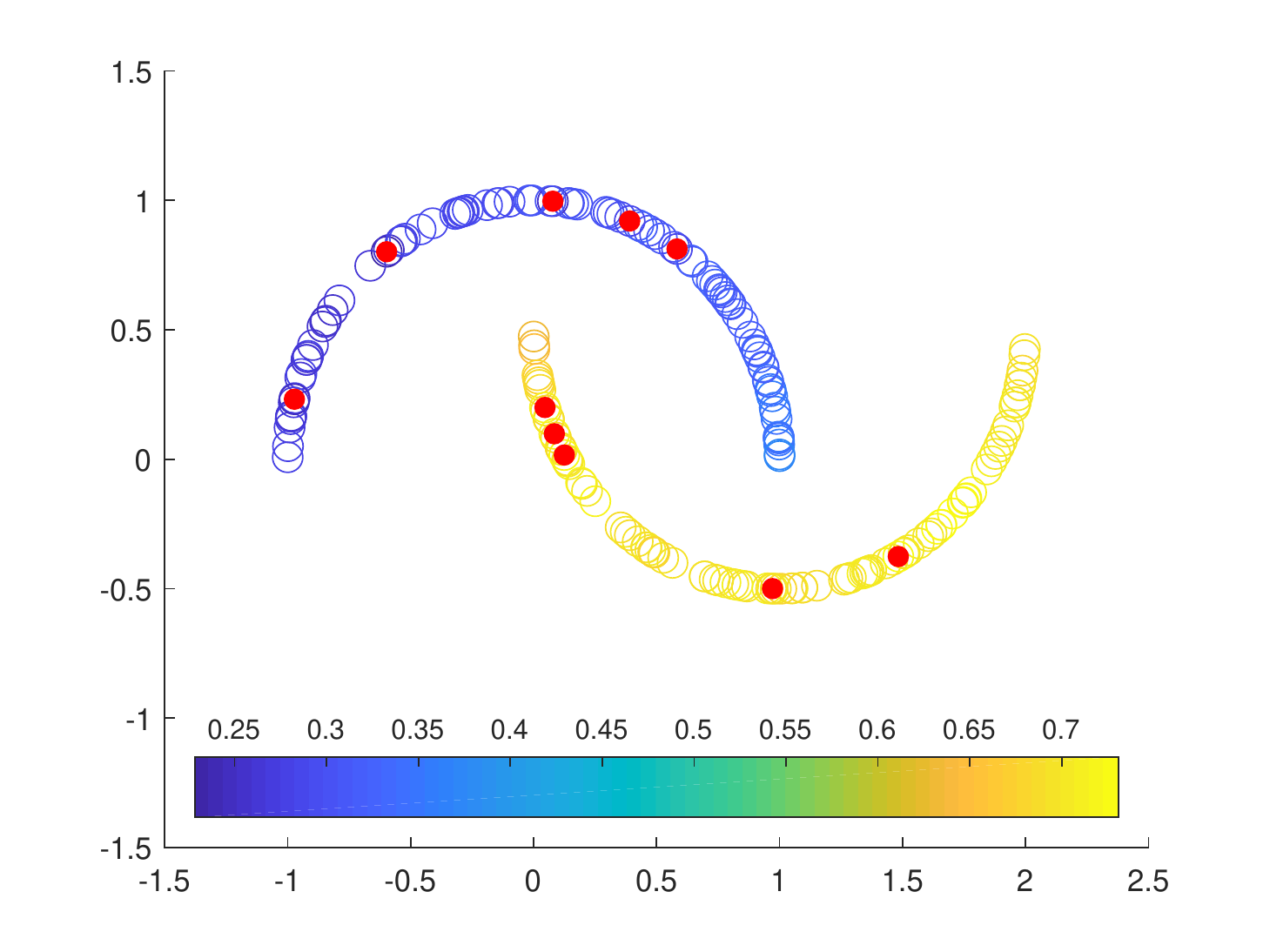}
		\centering
		\includegraphics[width=1\linewidth]{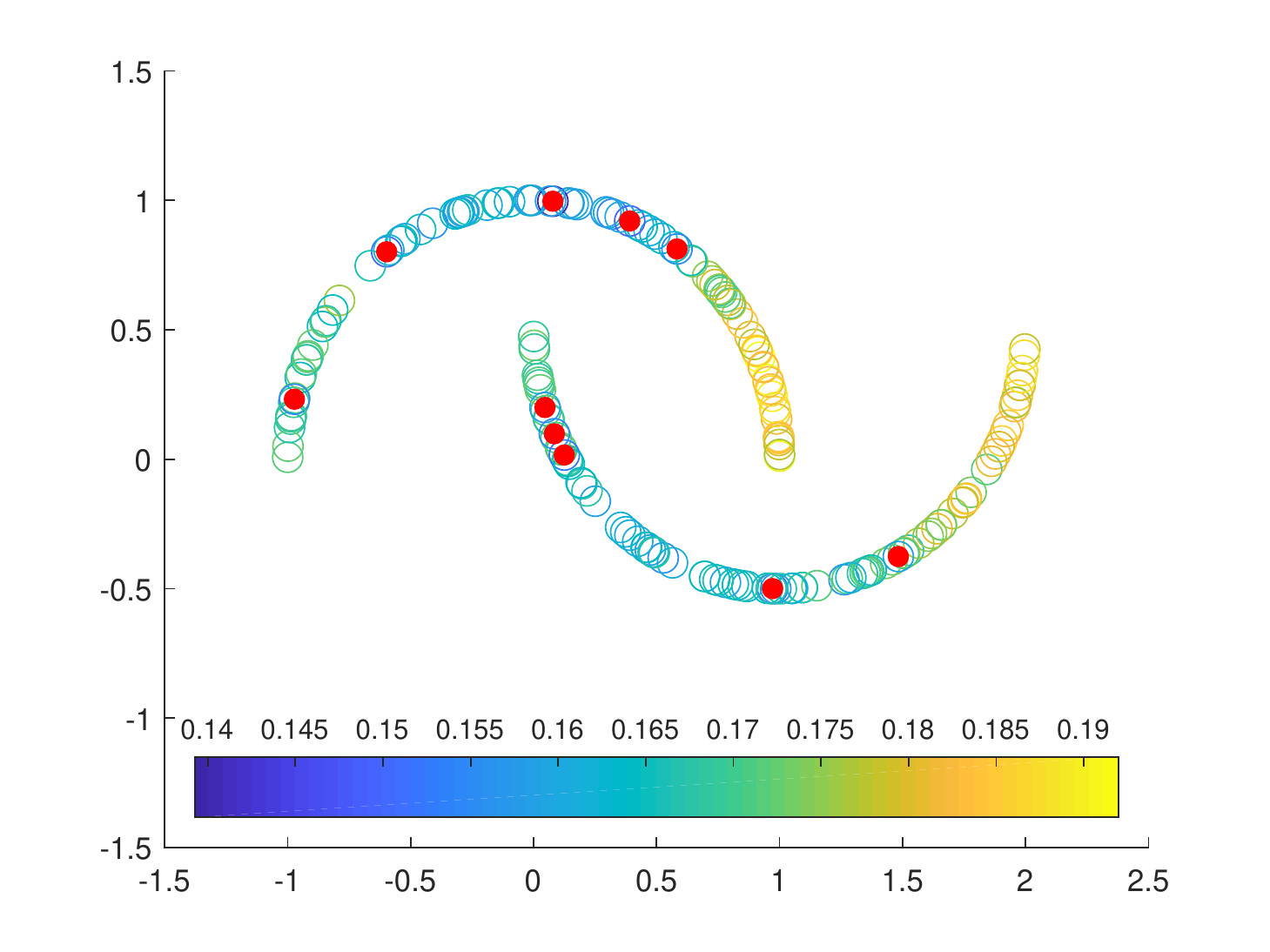}
		\caption{Posterior inference for a synthetic dataset. Each circle represents a feature. Labeled features are marked with red dots. Upper: posterior mean of probabilities of cluster assignment. Lower: the corresponding posterior standard deviations.}
		\label{figure:two moons}
	\end{figure}

 	\paragraph{Illustrative Example}
 	Figure \ref{figure:two moons} contains a synthetic binary classification toy example where the features are sampled from two disjoint semi-circles, corresponding to two classes. We are given $N = 200$ features, with only $n = 10$ of them labeled (marked with red dots), and aim to classify the unlabeled ones. Using a  graph-based prior defined following the ideas in Section \ref{sec:prior}, we compute the posterior mean ---which estimates the probability with which each data point belongs to the lower semi-circle---  and the posterior standard deviation ---which represents the uncertainty in the estimation. Thresholding the posterior mean at $0.5$ for classification, the results indicate very high accuracy, with lower uncertainty in the reconstruction near labeled features.  This example demonstrates a typical scenario in SSL, where the geometric information carried by the unlabeled data helps to achieve good learning performance with few labels.

 	\paragraph{Outline}
 	Having introduced the graph-based Bayesian formulation of SSL, we are ready to outline the questions that will be explored in the rest of this article, along with their practical motivation:
 	\begin{enumerate}
 		\item {\bf Prior Design:} How to define the latent field $u_N$ so that the prior $\pi_N$ promotes smoothness, facilitates computationally efficient inference, and yields a posterior that achieves optimal estimation for a large class of labeling functions?
 		\item {\bf Posterior Continuum Limit:}  For  a fixed number $n$ of labels, do the graph-based posteriors $\mu_N$ converge to a well-defined probability measure in the limit of  a  large number $N$ of features?
 		\item {\bf Posterior Sampling:}  Can  we design sampling algorithms whose rate of convergence does not deteriorate in the large $N$ limit?
 		\item {\bf Posterior Contraction:}  In  the joint limit where both $N$ and $n$ are allowed to grow, does the posterior distribution concentrate around the true labeling function? How should $N$ scale with $n$ to achieve optimal estimation? 
 	\end{enumerate}
 	These questions, along with their interrelations, will be considered in the next four sections. We will focus on graph-based Bayesian SSL, but related asymptotic analyses of SSL include \cite{nadler2009semi,bertozzi2021posterior}. 
 	An overarching theme in our Bayesian setting will be to guarantee that the prior and the posterior distributions are well defined in the limit of  a  large number of features (interpreted as graph nodes).  This idea is formalized through the study of \emph{continuum limits}, that play an essential role in  understanding the statistical performance of graph-based Bayesian SSL and the scalability of sampling algorithms.
 	
 	In order to set the theory on a rigorous footing, we will adopt the \emph{manifold assumption} that the features lie on a hidden low dimensional manifold \cite{belkin2004semi} embedded in an Euclidean space; for the study of posterior contraction, $f_0$ will be assumed to be a smooth function defined in this manifold. We emphasize that the manifold setting is used only for theoretical insight, but the methodology is applicable beyond this setting. The manifold assumption is widely adopted in machine learning and high dimensional statistics, and encapsulates the empirical observation that high dimensional features often exhibit low dimensional structure. 
 	
 	 We end this section by showing, in a concrete application, the interpretation of features and labels, as well as the intuition behind manifold and smoothness assumptions. The MNIST dataset $\{x_i\}_{i=1}^N$ consists of $N=60000$ images of hand-written digits from 0 to 9. We may want to classify images given labels $\{y_i\}_{i=1}^n$ with $y_i \in \{0, \ldots, 9\}$ and $n \ll N.$  Each image $x_i \in \mathbb{R}^d$ is a $d = 784$ dimensional vector, but the space of digits has been estimated \cite{hein2005intrinsic} to have dimension around $m = 10,$  and can be conceptualized as an $m$ dimensional manifold embedded in $\mathbb{R}^d$. The smoothness assumption reflects the idea that images that are similar are likely to correspond to the same digit, and should therefore receive the same label. As in the synthetic example of Figure \ref{figure:two moons} we need to construct a suitable prior for functions over the features $x_i$ and study posterior sampling algorithms.

 	\section{Prior Design}\label{sec:prior}
 	In this section we discuss the definition of the Gaussian random vector $u_N$ used to specify the prior $\pi_N.$ 
 	It will be convenient to think of $u_N$ as a random function over $\mathcal{M}_N:=\{x_1,\ldots,x_N\}$, or a random process discretely indexed by $\mathcal{M}_N$. We will denote by $u_N(i) := u_N(x_i)$ the value of the $i$th coordinate of $u_N$. Such notation will help highlight the analogies between the design of our discretely indexed random vector $u_N$ and the design of Gaussian processes (GPs) in Euclidean domains.

 	  In GP methodology, it is important to impose adequate smoothness assumptions. For instance, in the popular Mat\'ern class of GPs (to be defined shortly in Section \ref{sec:MGP})  in Euclidean space, the mean square differentiability of sample paths is tuned by a smoothness parameter. However, here we seek to define a discretely indexed random vector over abstract features---not necessarily embedded in Euclidean space---and the usual notions of smoothness are not readily applicable. To circumvent this issue, we will rely on a matrix $W$ of pairwise similarities between features. At a high level, we would like $u_N$ to be a random function that varies smoothly over $\mathcal{M}_N$ with respect to the pairwise similarities, i.e., the function values $u_N(i)$ and $u_N(j)$ for $i\neq j$ should be close if the similarity $W_{ij}$ between $x_i$ and $x_j$ is high. In other words, if we view the features $\mathcal{M}_N$ as a graph whose edge information is encoded in the similarity matrix $W$, then we wish $u_N$ to be regular with respect to the graph structure of $(\mathcal{M}_N,W)$. Techniques from spectral graph theory will allow us to construct a random vector that fulfills this smoothness requirement. 
 	
 	\subsection{GPs Over Graphs}
\begin{figure*}[!htb]
		\centering
		\minipage{0.249\textwidth}
		\includegraphics[width=\linewidth]{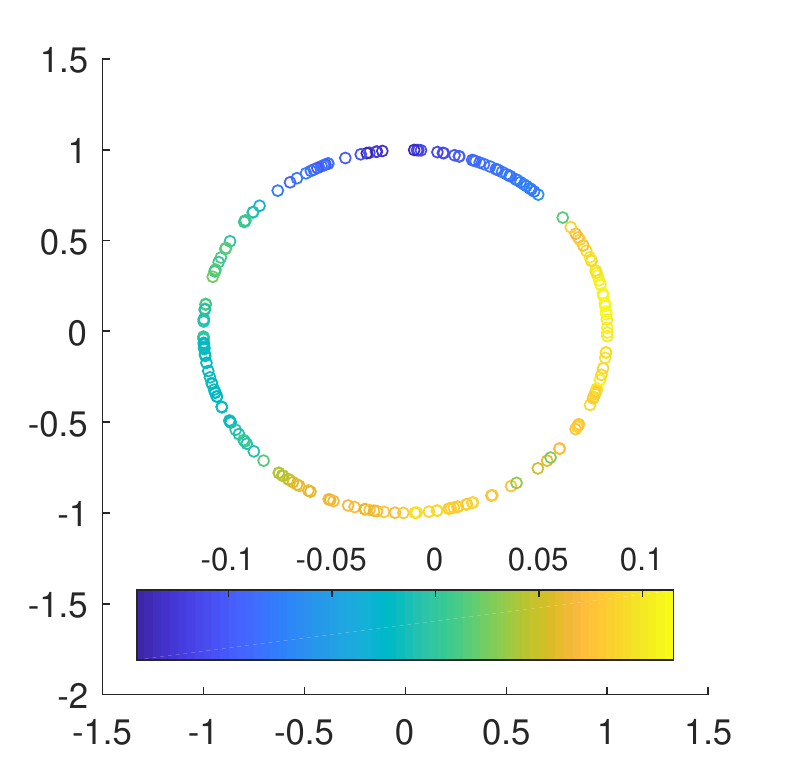} 
		\endminipage
		\minipage{0.249\textwidth}
		\includegraphics[width=\linewidth]{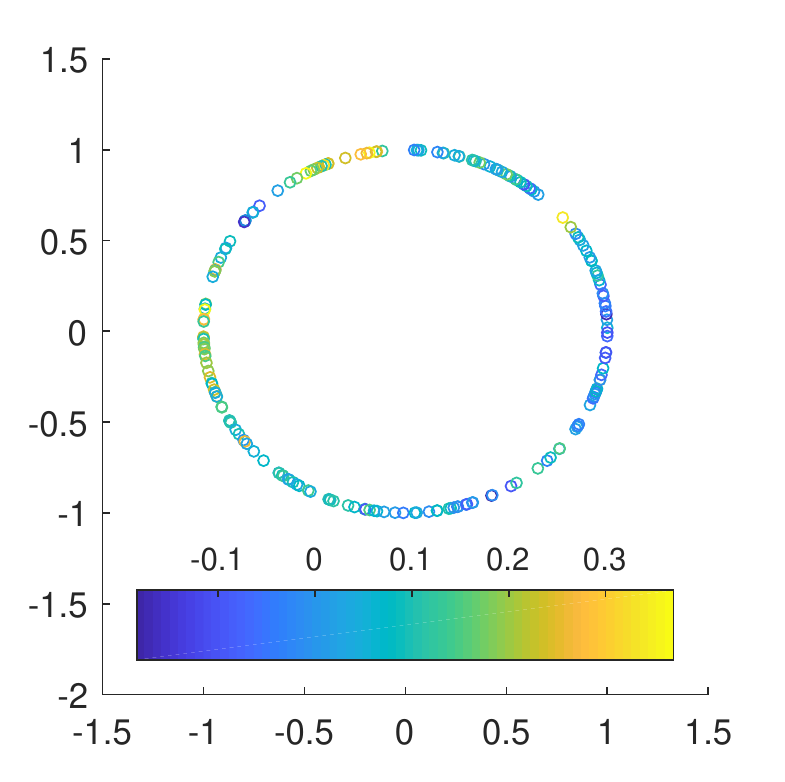}
		\endminipage
		\minipage{0.249\textwidth}
		\includegraphics[width=\linewidth]{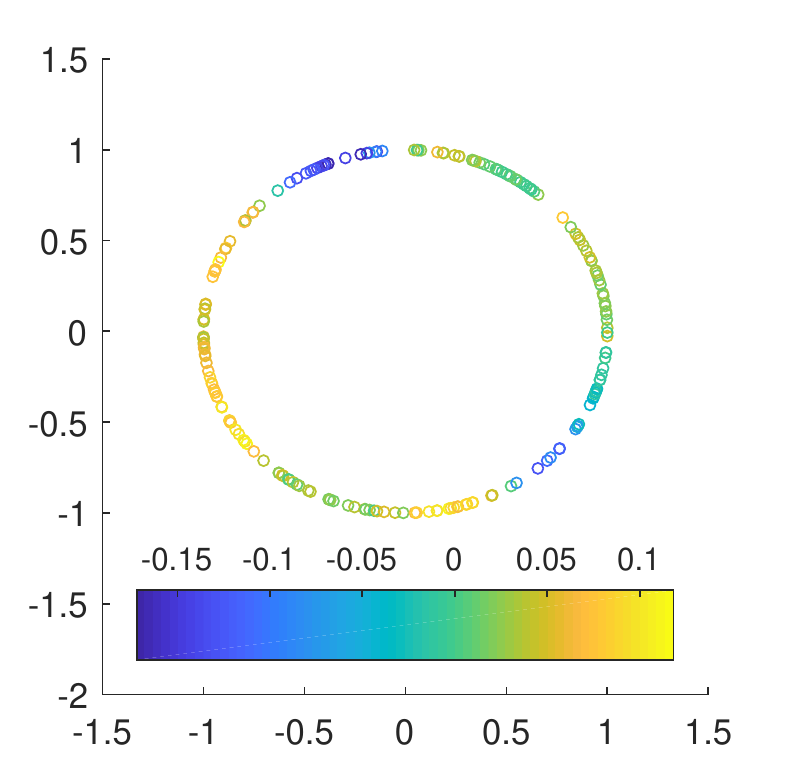}
		\endminipage
		\minipage{0.249\textwidth}
		\includegraphics[width=\linewidth]{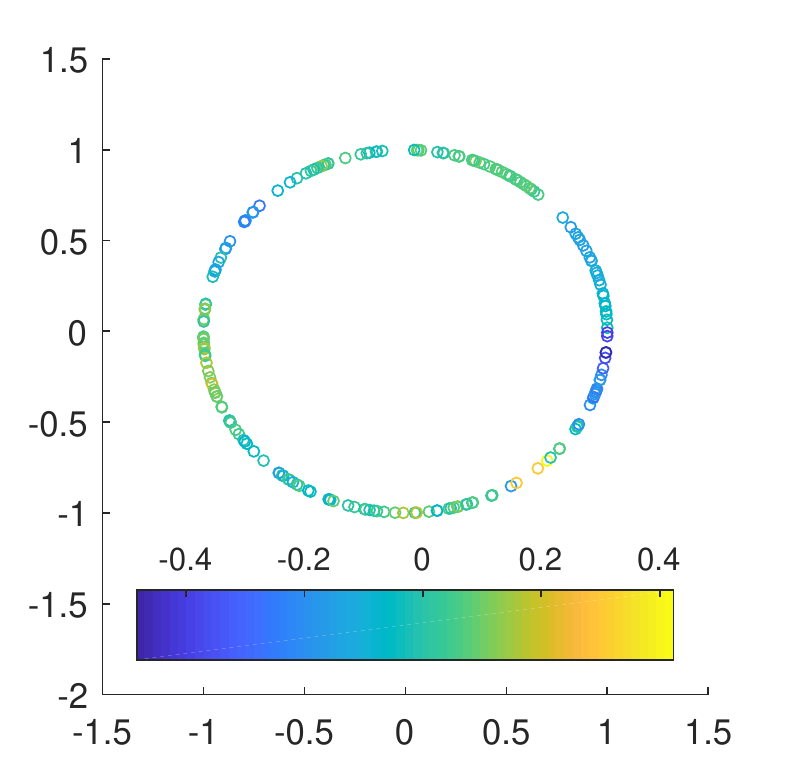}
		\endminipage\hfill
		\centering
		\minipage{0.249\textwidth}
		\includegraphics[width=\linewidth]{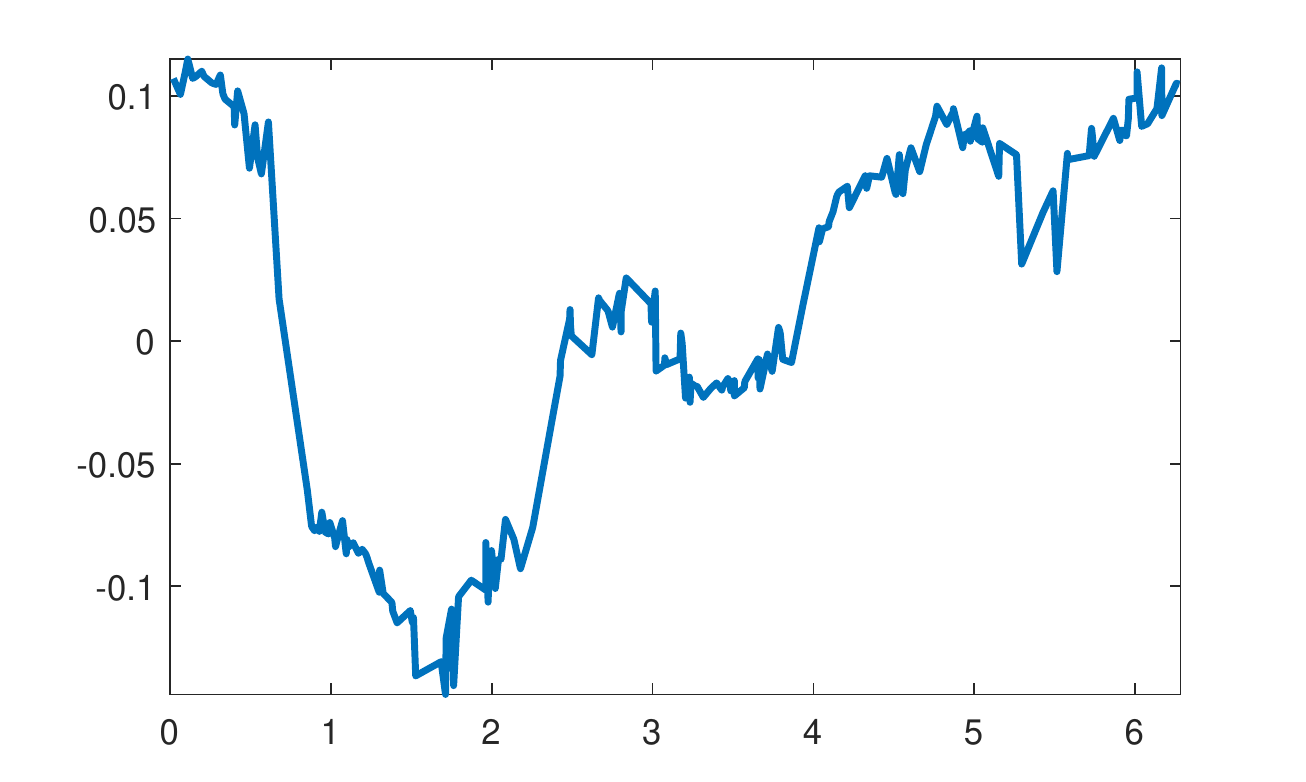} 
		\subcaption{$\tau=1, s=2$}
		\label{fig:sample1}
		\endminipage
		\minipage{0.249\textwidth}
		\includegraphics[width=\linewidth]{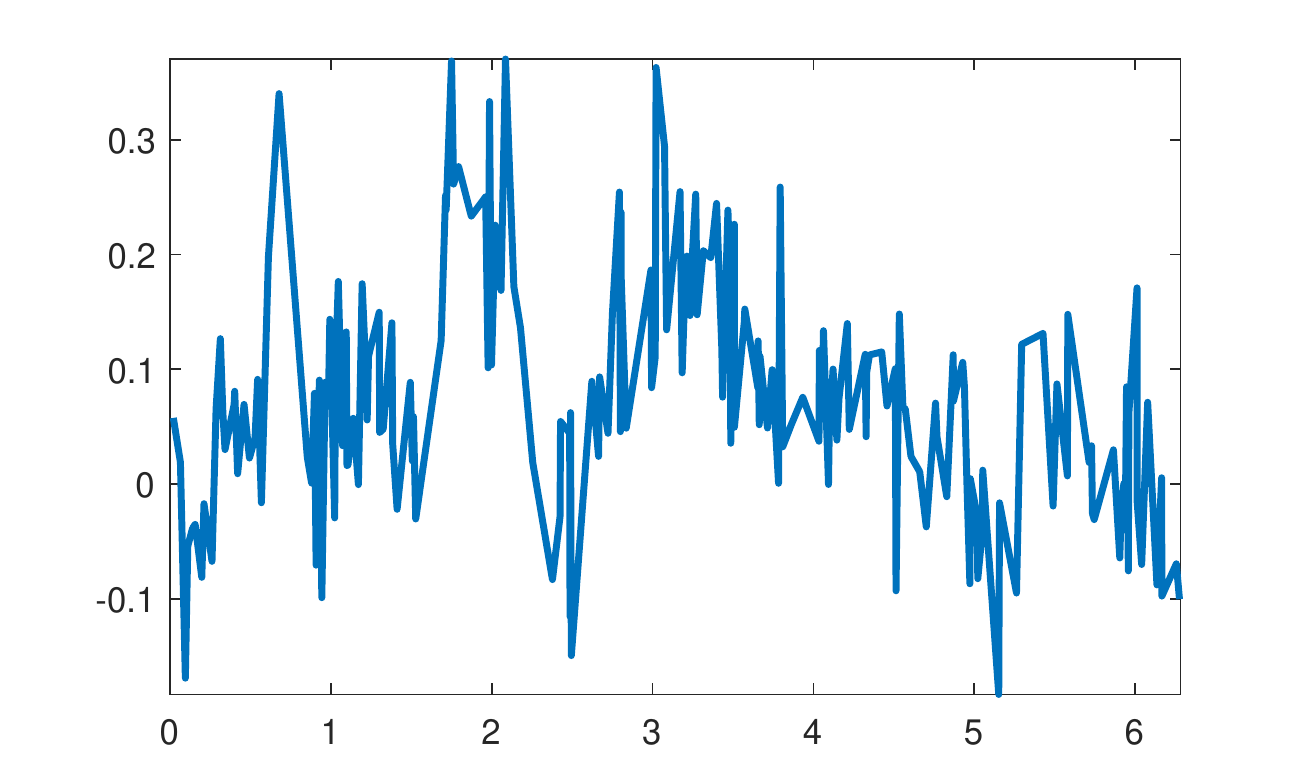}
		\subcaption{$\tau=30, s=2$}
		\label{fig:sample2}
		\endminipage
		\minipage{0.249\textwidth}
		\includegraphics[width=\linewidth]{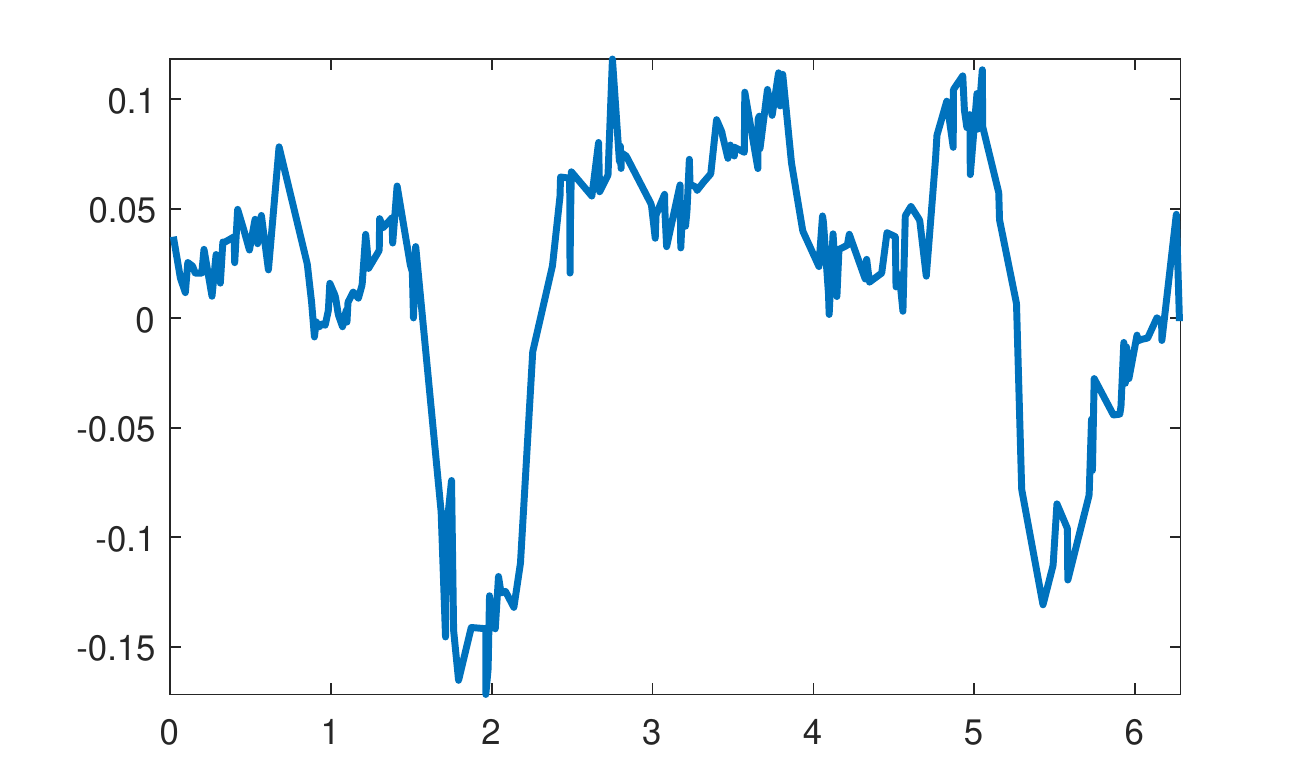}
		\subcaption{$\tau=30, s=4$}
		\label{fig:sample3}
		\endminipage
		\minipage{0.249\textwidth}
		\includegraphics[width=\linewidth]{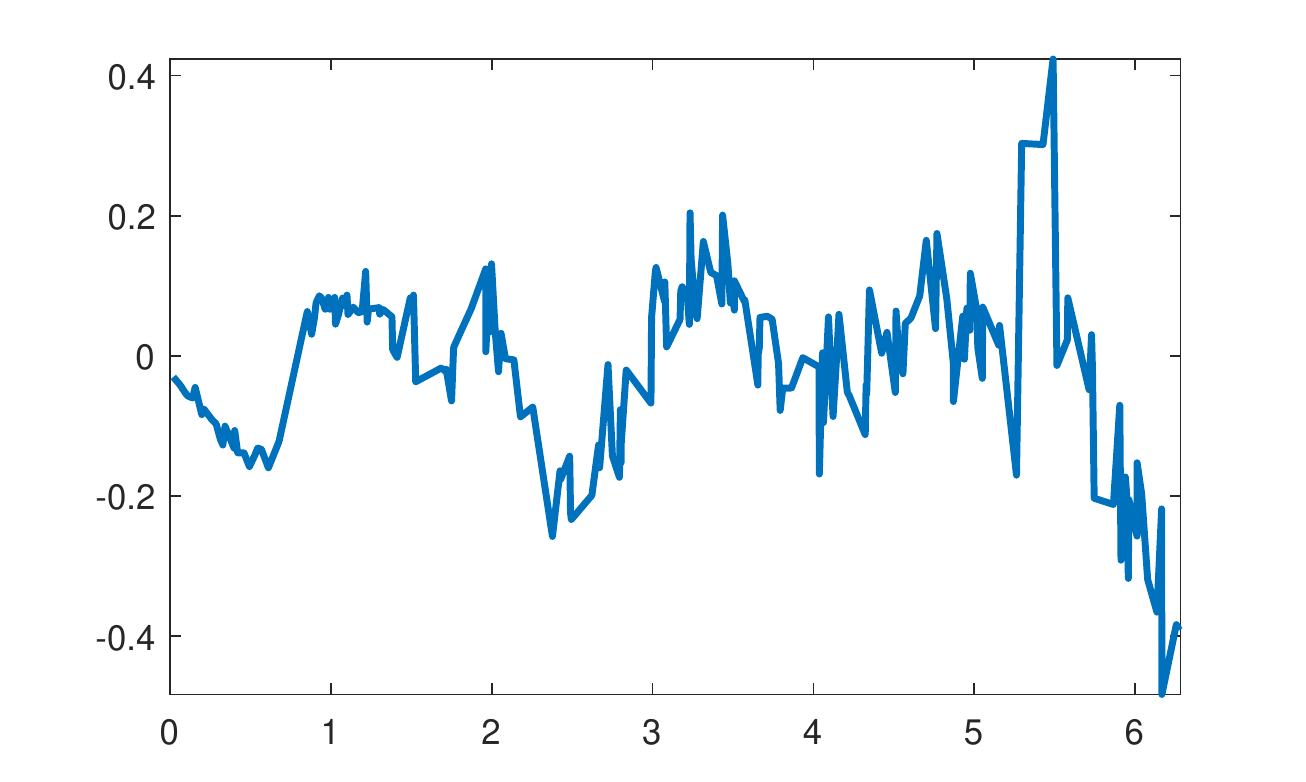}
		\subcaption{$\tau$ varying, $s=2$}
		\label{fig:sample varying}
		\endminipage
		\caption{Plots of samples of \eqref{eq:stationary graph prior} for different $\tau$'s and $s$'s when the $x_i$'s are sampled from the unit circle. The second row unfolds the plots in the first to the interval $[0,2\pi]$ for better visualization of the fluctuations.}
		\label{fig:my_label}
	\end{figure*}

    Graph-Laplacians, reviewed here succinctly, will be central to our construction. Given the similarity matrix $W\in\mathbb{R}^{N\times N}$, let $D\in\mathbb{R}^{N\times N}$ be the diagonal matrix with entries $D_{ii}=\sum_{j=1}^N W_{ij}$. The \emph{unnormalized} graph-Laplacian is then the matrix $\Delta_N=D-W$. Several normalized graph-Laplacians can also be considered (see e.g. \cite{von2007tutorial}), but for our purpose we will focus on the unnormalized one. From the relation
 	\begin{align}\label{eq:GL quadratic form}
 	v^T\Delta_N v =\frac12 \sum_{i,j=1}^N W_{ij}|v(i)-v(j)|^2, \quad v\in\mathbb{R}^N
 	\end{align}
 	we readily see that $\Delta_N$ is positive semidefinite. Moreover, \eqref{eq:GL quadratic form} implies that if we identify $v$ with a function over $\mathcal{M}_N$, then $v$'s that change slowly with respect to the similarities lead to smaller values of $v^T\Delta_Nv$. This observation suggests considering Gaussian distributions of the form $\mathcal{N}(0,\Delta_N^{-1})$ since its negative log density is proportional to \eqref{eq:GL quadratic form} (up to an additive constant) and therefore favors those ``smooth'' $v$'s. However $\Delta_N$ is singular, and the above Gaussian would be degenerate. To remedy this issue, and to further exploit the regularizing power of $\Delta_N$, we will instead consider 
 	\begin{align}\label{eq:stationary graph prior}
 	u_N\sim \mathcal{N} \bigl(0,(\tau I_N+\Delta_N)^{-s} \bigr)
 	\end{align}
 	with $\tau,s>0$ and $I_N$ the identity matrix. Here we have two additional parameters $\tau$ and $s$ which enhance the modeling flexibility. Roughly speaking, $\tau$ and $s$ control, respectively, the inverse lengthscale and smoothness of the samples (interpreted as functions over the graph).  
 	To see this, we can write the Karhunen-Lo\`eve expansion of $u_N$ in \eqref{eq:stationary graph prior} 
 	\begin{align}
 	u_N=\sum_{i=1}^N (\tau+\lambda_{N,i})^{-s/2}\xi_i \psi_{N,i} \quad \xi_i\overset{\text{i.i.d.}}{\sim}\mathcal{N}(0,1), \label{eq:KL graph prior}
 	\end{align}
 	where $\{(\lambda_{N,i},\psi_{N,i})\}_{i=1}^n$ are the eigenvalue -- eigenvector pairs of $\Delta_N$ with increasingly ordered eigenvalues. The eigenvectors become more oscillatory as $i$ increases, and therefore a larger $s$ ---which implies faster decay of the coefficients--- yields more regular sample paths, whereas a larger $\tau$ incorporates more essential frequencies and makes the sample paths more oscillatory. Figures \ref{fig:sample1}, \ref{fig:sample2} and \ref{fig:sample3} demonstrate this behavior for three sets of parameters when the $x_i$'s are sampled from the unit circle. Finally, to further enhance the modeling flexibility, one can replace $\tau I_N$ with a vector $\tau_N$ with positive entries. Doing so introduces a form of nonstationary local behavior, as shown in Figure \ref{fig:sample varying}, where $\tau_N$ increases from 1 (the left end) to 30 (the right end).

 	\subsection{Connection With Mat\'ern GP}\label{sec:MGP}
 	Besides the regularizing effect of the graph-Laplacian described above, the Gaussian distribution \eqref{eq:stationary graph prior} is also motivated by a  close connection to Mat\'ern GPs on Euclidean spaces. To start with, recall that the Mat\'ern covariance function takes the following form
 	\begin{align}
 	c(x,x')=\sigma^2 \frac{2^{1-\nu}}{\Gamma(\nu)}\left(\kappa|x-x'|\right)^{\nu}K_{\nu}\left(\kappa|x- x'|\right),
 	\end{align}
 	for $x,x'\in\mathbb{R}^{d}$. Here $\Gamma$ is the gamma function and $K_{\nu}$ is the modified Bessel function of the second kind. The Mat\'ern GP is a GP with the Mat\'ern covariance function. It is a popular modeling choice in Bayesian methodology due to the flexibility offered by the three parameters $\sigma,\nu,\kappa$ that control, respectively, the marginal variance, sample path smoothness, and correlation lengthscale. As we will see shortly, it turns out that we can view the finite dimensional Gaussian \eqref{eq:stationary graph prior} as a discrete analog of the Mat\'ern GP where the parameters $\nu$ and $\kappa$ play similar roles as our $\tau$ and $s$. 
 	
 	The key connection is the stochastic partial differential equation (SPDE) representation of Mat\'ern GP proved by \cite{whittle1963stochastic}, which says that the Mat\'ern GP $u$ is the unique stationary solution to  
 	\begin{align}
 	(\kappa^2-\Delta)^{\nu/2+d/4}u=\sigma \sqrt{\frac{(4\pi)^{d/2}\Gamma(\nu+d/2)\kappa^{2\nu}}{\Gamma(\nu)}}\mathcal{W}, \label{eq:SPDE}
 	\end{align}
 	where $\Delta$ is the usual Laplacian and $\mathcal{W}$ is a spatial white noise with unit variance. With this in mind, we can rewrite  \eqref{eq:stationary graph prior} in a similar fashion as 
 	\begin{align}
 	(\tau I_N+\Delta_N)^{s/2} u_N = \mathcal{W}_N, \label{eq:graph SPDE}
 	\end{align}
 	where $\mathcal{W}_N\sim \mathcal{N}(0,I_N)$. Now, ignoring the marginal variance in \eqref{eq:SPDE}, one can immediately see \eqref{eq:graph SPDE} as a discrete analog of \eqref{eq:SPDE} under the relation $s=\nu+d/2$ and $\tau=\kappa^2$. In other words, we can interpret \eqref{eq:stationary graph prior} as a Mat\'ern GP over the graph $(\mathcal{M}_N,W)$. 
 	
 	\subsection{Prior Continuum Limit}\label{ssec:priorcontinuumlimit}
 	If we impose certain assumptions on the graph $(\mathcal{M}_N,W)$, it can be shown that our graph Mat\'ern GP $u_N$ is not only a discrete analog of the usual Mat\'ern GP, but a consistent approximation of certain continuum Mat\'ern-type GPs. To formalize this statment, we rely on the manifold assumption that we had previously foreshadowed. Suppose now that the $x_i$'s are independently sampled from the uniform distribution in the manifold $\mathcal{M}$. We then have the following result (see \cite[Theorem 4.2 (1)]{garcia2018continuum} and \cite[Theorem 4.2]{sanz2020spde} for the formal version):
 	\begin{Result}
 		Under a manifold assumption, the graph Mat\'ern GP \eqref{eq:stationary graph prior} converges to a Mat\'ern-type GP on $\mathcal{M}$ provided that the similarity $W$ is suitably defined and the  
 		smoothness parameter $s$ is sufficiently large. 
 	\end{Result}
 	We next provide some further context for this result. First, the limiting Mat\'ern-type GP on $\mathcal{M}$ is defined by
 	\begin{align}
 	u\sim \mathcal{N} \bigl(0,(\tau I-\Delta_{\mathcal{M}})^{-s}\bigr), \label{eq:MGP manifold}
 	\end{align}
 	where $I$ is the identity and $\Delta_{\mathcal{M}}$ is the Laplace-Beltrami operator (the manifold analog of the usual Laplacian) on $\mathcal{M}$. By convention, $\Delta_{\mathcal{M}}$ is a negative semidefinite operator, which explains the minus sign. Just as the connection between \eqref{eq:stationary graph prior} and the SPDE representation of Mat\'ern GP, we can see \eqref{eq:MGP manifold} as a manifold analog of Mat\'ern GP defined by lifting \eqref{eq:SPDE}. In particular, we can write a similar series representation of \eqref{eq:MGP manifold} 
 	\begin{align}
 	u=\sum_{i=1}^{\infty} (\tau+\lambda_i)^{-s/2}\xi_i\psi_i,\quad \xi_i\overset{\text{i.i.d.}}{\sim}\mathcal{N}(0,1), \label{eq:KL continuum prior}
 	\end{align}
 	in terms of the eigenpairs $\{(\lambda_i,\psi_i)\}_{i=1}^{\infty}$ of $-\Delta_{\mathcal{M}}$. The eigenfunctions encode rich information about the geometry of $\mathcal{M}$ and form a natural basis of functions over $\mathcal{M}$. Comparing \eqref{eq:KL graph prior} and \eqref{eq:KL continuum prior}, it is reasonable to expect that large $N$ convergence will hold provided that we have convergence of the corresponding eigenvalues and eigenfunctions. To achieve this, we need to carefully construct the similarity matrix $W$ so that the graph-Laplacian $\Delta_N$ is a good approximation of $-\Delta_{\mathcal{M}}$. If we assume that $\mathcal{M}$ is an $m$-dimensional compact submanifold of $\mathbb{R}^d$, then this is indeed the case if we set 
 	\begin{align}
 	W_{ij}=\frac{2(m+2)}{N\nu_m h_N^{m+2}}\mathbf{1}\{|x_i-x_j|<h_N\}, \label{eq:graph weights}
 	\end{align}
 	where $\nu_m$ is the volume of the $m$-dimensional unit ball and $h_N$ is a user chosen graph connectivity parameter satisfying 
 	\begin{align}
 	\frac{(\log N)^{c_m}}{N^{1/m}}\ll h_N \ll \frac{1}{N^{1/2s}},\label{eq:graph connectivity}
 	\end{align}
 	with $c_m=3/4$ if $m=2$ and $c_m=1/m$ otherwise.
 	Small values of $h_N$ induce sparse graphs, which are easier to work and compute with; see Section \ref{sec:sparsity} below. However, very small values of $h_N$ render graphs that are so weakly connected that they cannot induce any level of smoothness in the functions that are likely to be generated by the prior $\pi_N$. It is thus important to set the connectivity $h_N$ appropriately in order to take advantage of sparsity while at the same time recovering the geometric information of $\M$. The specific lower bound in \eqref{eq:graph connectivity} characterizes the level of resolution of the implicit discretization of the manifold induced by the $x_i$'s. We require $h_N$ to be larger than this quantity to capture the geometry of the underlying manifold.  Under these conditions on $h_N$, it can be shown that $\Delta_N$ converges spectrally towards $-\Delta_\M$. Other types of graphs such as $k$-nearest neighbors and variable bandwidth graphs can also be employed, and recent work \cite{garcia2019variational} have shown spectral convergence in these settings.

 	\subsection{Sparsity}
 	\label{sec:sparsity}
 	So far we have discussed the construction of our prior from a modeling perspective, motivated by the regularizing power of the graph-Laplacian and the connection with usual Mat\'ern GPs. We close this subsection by mentioning its sparseness. Notice that with our choice of weights \eqref{eq:graph weights}, the similarity matrix $W$ ---and hence the graph-Laplacian $\Delta_N$--- are sparse. Indeed, one can show that for
 	\begin{align*}
 	h_N\asymp  \sqrt{\frac{(\log N)^{c_m}}{N^{1/m}}}
 	\end{align*}
  the number of nonzero entries of $\Delta_N$ is $O(N^{3/2})$. 
 	Therefore, for small integer $s$ in \eqref{eq:stationary graph prior}, we are left with a Gaussian with sparse precision matrix, and numerical linear algebra methods for sparse matrices can be employed to speed-up computation.  This is important for posterior inference algorithms that may require factorizing $\Delta_N$.  Similar conclusions can be reached with $k$-nearest neighbors graphs.

 	\section{Posterior Continuum Limit}\label{sec:postcontinuumlimit}
 	
 	In this section we discuss the convergence of the posterior $\mu_N$ for large $N$ (and fixed $n$) towards a \textit{continuum posterior} $\mu$ defined later on. For now, it suffices to note that the continuum posterior is naturally characterized as a probability distribution over the space $L^2(\M)$.
When formalizing a notion of convergence for posteriors, a challenge arises: the measures $\mu_N$ and $\mu$ are probability measures defined over different spaces, i.e. $L^2(\M_N)$ and $L^2(\M)$, respectively. In what sense should these measures be compared? In what follows, we present a possible solution to this question, which also arises in the rigorous analysis of continuum limits for prior distributions considered in the previous section.

 \subsection{Lifting to the space $\mathcal{P}(TL^2)$}

In order to compare the measures $\mu_N$ and $\mu$, we start by introducing a space where we can directly compare functions in $L^2(\M_N)$ with functions in $L^2(\M)$. We let $TL^2$ be the set: 
\begin{equation*}
 TL^2 := \bigl\{   (\theta, g) \: : \:   \theta \in \mathcal{P}(\mathcal{M}), \,\, g \in L^2(\M, \theta)      \bigr\} .
 \end{equation*}
 In words, $TL^2$ is the collection of pairs of the form $(\theta,g)$, where $\theta$ is a probability measure over $\M$ and $g$ is an element in $L^2(\M, \theta)$. For us, the most important choices for $\theta$ are the empirical measure associated to the samples $x_i$ and the data generating distribution $\law(X)$. We use the simplified notation $L^2(\M_N)$ and $L^2(\M)$ to denote the $L^2$ spaces for these two choices of $\theta$. $TL^2$ can be formally interpreted as a fiber bundle over the manifold $\mathcal{P}(M)$:  each $\theta \in \mathcal{P}(M)$ possesses a corresponding $L^2$ fiber. 
 
 We endow $TL^2$ with the following distance:
 \begin{align*}
 & d_{TL^2}\bigl((\theta_1,h_1),(\theta_2,h_2)\bigr)^2 := \\ & \inf_{\gamma \in \Gamma(\theta_1, \theta_2)}  \iint_{\mathcal{M} \times \mathcal{M}} \hspace{-0.75cm} \bigl(d_\mathcal{M}^2 (x,\tilde{x})   + | h_1(x) - h_2(\tilde{x})|^2 \bigr) \, d\gamma(x,\tilde{x}),
 \end{align*}
 where $\Gamma(\theta_1, \theta_2)$ represents the set of \textit{couplings} between $\theta_1$ and $\theta_2$ ---that is, the set of probability measures on $\M \times \M$ whose first and second marginals are $\theta_1$ and $\theta_2,$ respectively--- and $d_\M$ denotes the geodesic distance in $\M$. It is possible to show that the $ d_{TL^2}$ metric is, indeed, a distance function. Moreover, the topology induced by $d_{TL^2}$ in each fixed fiber $L^2(\M, \theta)$  coincides with the topology induced by the natural topology of the Hilbert space $L^2(\M, \theta)$, a fact that motivates the notation $TL^2$, which suggests an $L^2$-like convergence after transportation.  We refer to \cite{ThorpeTLp} for further details.

We proceed to define a notion of convergence for the posteriors $\mu_N$ as $N \rightarrow \infty$. As discussed above, the $TL^2$ space allows us to see $L^2(\M_N)$ and $L^2(\M)$ as subsets of the bigger common space $TL^2$. In turn, the measures $\pi$ and $\pi_N$, as well as the measures $\mu$ and $\mu_N$, can then be all interpreted as probability measures on the space $TL^2$. Using this ``lifting" we can now interpret the statement $\mu_N \rightarrow \mu$ as $N \rightarrow \infty$, as a statement about the weak convergence of probability measures in the metric space $TL^2$. Further properties of the space $TL^2$ allow us to use a collection of theorems, such as Portmanteau's and Prokhorov's, to characterize convergence and compactness in the space $\mathcal{P}(TL^2)$.

After specifying the notion of convergence of $\mu_N$ towards $\mu$, we can now present a result, rigorously stated in \cite{garcia2018continuum}.
 	\begin{Result}
 	\label{res:1}
 		Under a manifold assumption, the graph-based posterior $\mu_N$ converges to a continuum limit posterior $\mu$ over functions on $\mathcal{M}$, provided that the similarity $W$ is suitably defined and the smoothness parameter $s$ is sufficiently large. 
 	\end{Result}
 	Further context for this result will be given next.

 	\subsection{Convergence of Posteriors}
 Now that we have discussed the precise way in which we formalize the convergence of $\mu_N$ towards $\mu$, we proceed to characterize $\mu$ and describe the tools used to deduce this convergence. For ease of exposition, we focus on the regression setting. 
 
 First, we notice that the posterior distribution $\mu_N$, introduced in Section \ref{ssec:BayesvsOpt} via Bayes's formula, can be characterized variationally. Indeed, $\mu_N$ is the solution to the optimization problem
 	$$\mu_N = \underset{\nu_N}{\operatorname{arg\, min}}\,\, J_N(\nu_N),$$
 	where, 	for $ \nu_N \in \mathcal{P} \bigl(L^2(\mathcal{M}_N)\bigr),$
 	$$J_N(\nu_N) :=  \dkl(\nu_N \| \pi_N) \,+  \int_{L^2(\mathcal{M}_N)} \ell(f_N; y)\, d \nu_N(f_N).$$
 Here $\dkl$ denotes the Kullback-Leibler divergence and $\ell(f_N; y)$ denotes the negative log-likelihood. The first term in $J_N$ will be small if $\nu_N$ is close to the prior $\pi_N,$ while the second term will be small if $\nu_N$ gives significant mass to $f_N$'s that are consistent with the labeled data. Thererfore, the minimizer $\mu_N$ of $J_N$ represents a compromise between matching prior beliefs and matching the observed labels. Following this variational characterization, we define the continuum posterior $\mu$ in direct analogy with the graph setting:
 	$$\mu = \underset{\nu}{\operatorname{arg\,min}}\,\, J(\nu),$$ 	
 	where, for $\nu \in \mathcal{P} \bigl(L^2(\mathcal{M})\bigr),$
 	$$ J(\nu) := \dkl(\nu \| \pi) \,+  \int_{L^2(\mathcal{M})} \ell(f; y) \, d \nu(f).$$
 	 The energies $J_N$ and $J$ can be extended to $\mathcal{P}(TL^2)$ by setting them to be infinity outside the fibers $L^2(\M_N)$ and $L^2(\M)$, respectively. This extension is convenient so as to have a collection of functionals defined over a common space. The variational characterization opens the door to the use of  tools in the calculus of variations, which allow to prove the convergence of minimizers of variational problems.  Indeed, the following three statements together imply the convergence of the minimizer of $J_N$ towards the minimizer of $J$, that is, the desired convergence of posteriors.  
 	\begin{enumerate}
 		\item For every converging sequence $\nu _N \rightarrow \nu$ we have $\liminf_{N \rightarrow \infty} J_N(\nu_N) \geq J(\nu)$.
 		\item For every $\nu$ there exists a sequence $\{ \nu_N\}_{N=1}^{\infty}$ such that $\limsup_{N \rightarrow \infty} J_N(\nu_N) \leq J(\nu)$.
 		\item Every sequence $\{ \nu_N\}_{N=1}^{\infty}$ in $\mathcal{P}(TL^2)$ satisfying
 		\[ \sup_{N } J_N(\nu_N) < \infty \] 
 		is precompact.
 	\end{enumerate}
As it turns out, it is possible to prove that, under the assumptions of  Result \ref{res:1}, these three statements hold simultaneously with probability one. 
The structure of  $J_N$ and $J$ ---where prior and likelihood appear separately--- facilitates the analysis.
 The most delicate part is to compare the prior distributions $\pi_N$ and $\pi$, that is, the first terms of $J_N$ and $J$. To provide some further intuition, we recall that a random variable $u_N$ sampled from the discrete prior $\pi_N$ takes the form:
 	\begin{align*}
 	u_N=\sum_{i=1}^N (\tau+\lambda_{N,i})^{-s/2}\xi_i \psi_{N,i} \quad \xi_i\overset{\text{i.i.d.}}{\sim}\mathcal{N}(0,1),
 	\end{align*}	
 while a sample $u$ from the continuum prior $\pi$ takes the form:	
 	\begin{align*}
 	u=\sum_{i=1}^\infty (\tau+\lambda_{i})^{-s/2}\xi_i \psi_{i}.
 	\end{align*}
Here we use the same random variables $\xi_i$ in both $u$ and $u_N$, thereby \textit{coupling} the measures $\pi_N$ and $\pi$. It can be shown that the $TL^2$ distance between $\psi_{N,i} $ and $\psi_i$ can be controlled with very high probability for all $i$ up to some mode $b $ smaller than $N$. We can thus expect that the sum of the first $b$ terms in $u_N$ is close, in the $TL^2$ sense, to the sum of the first $b$ terms in $u$. For modes larger than $b$, on the other hand, it will not be possible to obtain decaying estimates for the distance between the corresponding discrete and continuum eigenfunctions. This is to be expected as the graph cannot resolve the geometry of the manifold $\M$ at lengthscale $\sim h$. To control the higher modes, we must use the fact that the terms $( \tau + \lambda_{i})^{-s/2}$ can be controlled by a factor of the form $\sim b^{-s/m}$, as it follows from the well known Weyl's principle describing the growth of eigenvalues of Laplace-Beltrami operators on compact manifolds. Here it is worth recalling our discussion in earlier sections regarding the level of regularity induced by higher value of $s$: a large enough value of $s$ can be used to control the contribution of high order modes. The above argument eventually leads to the following estimate:
\begin{align*}
  \min _{\gamma \in \Gamma(\pi_N, \pi) } \int_{TL^2} \int_{TL^2}  (d_{TL^2}(v_N, v))^2 d \gamma (v_N, v) 
  \\ \leq  \mathbb{E} \left[ d_{TL^2}( u_N, u)^2\right]  \rightarrow 0,
\end{align*}
as  $N \rightarrow \infty$. In other words, in the Wasserstein space over $TL^2$, the measure $\pi_N$ converges towards $\pi$ as $N \rightarrow \infty$, and thus, the convergence holds also in the weak sense, implying the convergence of the prior terms. With the convergence of priors in hand, the proofs of statements 1-2-3 reduce to a careful use of lower semi-continuity properties of the Kullback-Leibler divergence. We refer to \cite{garcia2018continuum} for further details, and describe next why establishing continuum limits for posterior distributions is important in the design of scalable algorithms for posterior sampling.

 	\section{Posterior Sampling}\label{sec:sampling}
 	As noted in Section \ref{ssec:BayesvsOpt}, the construction of point estimates and confidence intervals in Bayesian inference rests upon computing expectations with respect to the posterior distribution. For instance, finding the posterior mean, marginal variances, and quantiles requires  one  to compute $\mathbb{E}_{\mu_N}[h]$ for various test functions $h: \mathbb{R}^N \to \mathbb{R}.$
 	When the posterior is not tractable ---such as in SSL classification--- expectations can be approximated using sampling algorithms. The goal of this section is to show how the continuum limit of posteriors described in Section \ref{sec:postcontinuumlimit} can be exploited to design Markov chain Monte Carlo (MCMC) sampling algorithms with a rate of convergence that is independent of the number $N$ of features. Subsection \ref{ssec:MCMC} contains the necessary background on the Metropolis-Hastings MCMC algorithm. In Subsection \ref{ssec:graphpCN} we introduce the graph \emph{preconditioned Crank-Nicolson} (pCN) algorithm, a Metropolis-Hastings scheme that exploits the continuum limit to ensure scalability to large datasets. Finally, in Section \ref{ssec:spectralgap} we discuss how the large $N$ scalability of the graph pCN algorithm can be formalized through the notion of uniform spectral gaps.

 	\subsection{Metropolis-Hastings  Sampler }\label{ssec:MCMC}
 	Metropolis-Hastings MCMC is one of the most widely used algorithms in science and engineering, and is a cornerstone of computational Bayesian statistics. The basic idea is simple: for a given sample size $K$,  the  Metropolis-Hastings sampler  approximates 
 	\begin{equation}\label{eq:MCMCapprox}
 	\mathbb{E}_{\mu_N}[h] \approx \frac1K \sum_{k=0}^K h(f_N^{(k)}),
 	\end{equation}
 	where $\{f_N^{(k)}\}_{k=0}^K$ are samples from a Markov chain whose kernel $\pmh$ satisfies detailed balance with respect to $\mu_N,$ that is, 
 	\begin{equation}\label{eq:detailed balanace}
 	\mu_N(f) \, \pmh(f,g)  = \mu_N(g)\, \pmh(g,f), \quad \forall f, g. 
 	\end{equation}
 	The detailed balance condition \eqref{eq:detailed balanace} guarantees that $\mu_N$ is the stationary distribution of the Markov chain, and, consequently, $f_N^{(k)}$ will be approximately distributed as $\mu_N$ for large $k,$ under mild assumptions. 
 	
 	The Metropolis-Hastings algorithm is built upon an accept/reject mechanism that turns a given \emph{proposal kernel} into a Metropolis-Hastings  Markov kernel $\pmh$ that satisfies the desired detailed balance condition.
 	\begin{center}
 		\smartdiagramset{border color=none, 
 			set color list={black!40!,black!40!,black!40!},
 			back arrow disabled=true}
 		\smartdiagram[flow diagram:horizontal]{Proposal kernel, Accept/ \\ reject, Metropolis-Hastings kernel}
 	\end{center}
 	Given the $k$-th sample $f_N^{(k)}$, the $(k+1)$-th sample is obtained following a two-step process. 
 	First, a proposed move is sampled
 	$g_N^{(k)} \sim q(f_N^{(k)}, \cdot)$ from the given proposal kernel $q.$ Second, the proposed move is accepted  with probability  $a(f_N^{(k+1)}, g_N^{(k+1)})$
 	and rejected with probability $1 - a(f_N^{(k+1)}, g_N^{(k+1)}).$ If the move is accepted, one sets $f_N^{(k+1)} = g_N^{(k+1)}$; if rejected, $f_N^{(k+1)} = f_N^{(k)}.$  
 	The Metropolis-Hastings acceptance probability 
 	\begin{equation*}
 	a(f, g) := \min \Bigl\{  1, \frac{\mu_N(g)}{\mu_N (f )} \frac{q(g,f)}{q(f,g)}  \Bigr\}
 	\end{equation*}
 	is defined in such a way that the procedure renders a Markov chain whose kernel $\pmh$ satisfies \eqref{eq:detailed balanace}. Moreover,  under mild assumptions the distribution $\mu_N^{(k)}$ of the $k$-th sample $f_N^{(k)}$  converges to $\mu_N$ as $k\to \infty$. How fast this convergence occurs ---and, as a consequence, how accurate the approximation \eqref{eq:MCMCapprox} is for a given sample size $K$--- depends crucially on the choice of proposal kernel $q$.   In the following subsection we introduce the graph pCN algorithm:  a Metropolis-Hastings MCMC algorithm that uses a specific proposal kernel to ensure that the rate of convergence of the chain $\mu_N^{(k)}$ to the posterior $\mu_N$ does not deteriorate in the large $N$ limit. 
 	
 	\subsection{The Graph pCN Algorithm}\label{ssec:graphpCN}
 	The proposal kernel $\qpcn$ of the graph pCN algorithm \cite{bertozzi2018uncertainty} is chosen so that it satisfies detailed balance with respect to the \emph{prior} distribution $\pi_N.$ For ease of exposition, we present the algorithm in the regression setting. Let $\vartheta \in (0,1)$ be a tuning parameter, and set
 	\begin{equation}\label{eq:pcnproposal}
 	g_N^{(k)} = (1-\vartheta^2)^{1/2} f_N^{(k)} + \vartheta \, \xi_N^{(k)}, \, \xi_N^{(k)} \sim \pi_N,
 	\end{equation}
 	where $\pi_N$ is the prior on $f_N$ introduced in Section \ref{sec:prior}, with covariance $C_N.$
 	A direct calculation shows that the Markov kernel   $$\qpcn(f,\cdot)=\mathcal{N} \bigl((1-\vartheta^2)^{1/2} f ,  \vartheta^2 C_N\bigr)$$
 	implicitly defined by the proposal mechanism \eqref{eq:pcnproposal} satisfies detailed balance with respect to $\pi_N.$
 	Therefore, for the graph-pCN algorithm, the Metropolis-Hastings acceptance probability is given by
 	\begin{align*}
 	\apcn(f,g) &= \min \Bigl\{  1, \frac{\mu_N(g)}{\mu_N (f )} \frac{\qpcn(g,f)}{\qpcn(f,g)}  \Bigr\} \\
 	&= \min \Bigl\{  1, \frac{ L(g;y) \pi_N(g)}{ L(f;y) \pi_N (f )} \frac{\qpcn(g,f)}{\qpcn(f,g)}  \Bigr\}  \\
 	&= \min \biggl\{  1, \frac{ L(g;y) }{ L(f;y) } \biggr\} ,
 	\end{align*}
 	where we used detailed balance of $\qpcn$ with respect to $\pi_N$ in the last equation. Note that the probability of accepting a move is hence completely determined by the value of the likelihood at the proposed move relative to its value at the current state of the chain. In particular, moves that lead to a higher likelihood are always accepted. Putting everything together, the graph pCN algorithm \cite{bertozzi2018uncertainty,trillos2017consistency} is outlined in Algorithm \ref{pCN-RRW. }.
 	\begin{algorithm}
 		\caption{Graph pCN}\label{pCN-RRW. }
 		\begin{algorithmic}
 			\BState {\bf Input:} Prior $\pi_N,$  likelihood $L(\cdot; y),$  $\vartheta \in (0,1).$ 
 			\BState {\bf Initialize:} Pick $f_N^{(0)}.$
 			\BState {\bf For} $k = 0, 1, 2, \ldots$ {\bf do:}
 			\begin{enumerate}
 				\item {\bf Proposal step:} 
 				Set $$g_N^{(k)} = (1-\vartheta^2)^{1/2} f_N^{(k)} + \vartheta \, \xi_N^{(k)}, \, \xi_N^{(k)} \sim \pi_N.$$
 				\vspace{-0.5cm} \item {\bf Accept/reject step:} 
 				Compute $$a(g_N^{(k)} , f_N^{(k)}) := \min \biggl\{1, \, \frac{L(g_N^{(k)} ;y)}{ L(f_N^{(k)}; y) } \biggr\}$$
 				and set $$ f_N^{(k+1)} := \begin{cases}
 				g_N^{(k)} \quad \quad \, \text{w.p.} \,\,\,\,\, a\bigl(g_N^{(k)} , f_N^{(k)}\bigr),   \\
 				f_N^{(k)} \quad \quad \text{w.p.} \,\,\,\,\, 1 -  a\bigl(g_N^{(k)} , f_N^{(k)}\bigr). 
 				\end{cases}$$
 				\item $k\to k+1.$
 			\end{enumerate}
 			\BState {\bf Output:} $f_N^{(k)}, k = 0, 1, \ldots$
 		\end{algorithmic}
 	\end{algorithm}
 	
 	Notice that the prior distribution ---and hence the unlabeled features--- are only used in the proposal step, while the likelihood function ---and hence the labels--- are only used in the accept/reject step. 
 	Therefore, one would expect that the acceptance rate should not fundamentally depend on the number of unlabeled features, provided that the prior approaches a continuum limit and the number of labels are kept fixed.  This insight can be formalized into a rigorous guarantee of algorithmic scalability,  discussed next.

 	\subsection{Uniform Spectral Gap}\label{ssec:spectralgap}
 	As noted above, under mild assumptions on the likelihood function, it is possible to show that the distribution $\mu_N^{(k)}$ of the $k$-th sample $f_N^{(k)}$ of the pCN algorithm converges to $\mu_N$ in the large $k$ limit. More precisely, for a suitable distance $d$ between probability measures, one can show that there are constants $c>0$ and $\epsilon_N \in (0,1)$ such that 
 	\begin{equation*}
 	d( \mu_N^{(k)} , \mu_N) \le c\, (1 - \epsilon_N)^k, \quad k  = 0, 1, \ldots
 	\end{equation*}
 	The  largest  $\epsilon_N$ satisfying this requirement is called the \emph{spectral gap} of the chain. A  large  spectral gap implies fast convergence of the chain. 
 	In particular, a positive spectral gap is sufficient to ensure the consistency and asymptotic normality of the estimator  \eqref{eq:MCMCapprox} for suitable test functions. It is therefore important to understand if the spectral gaps $\epsilon_N$ deteriorate  (i.e. decay to zero) as $N$ grows. The following result, formalized in  \cite{trillos2017consistency},  indicates that the spectral gaps for the graph pCN algorithm are \emph{uniform}, meaning that they are bounded from below by a positive constant independent of $N$. 
 	\begin{Result}
 		Under the conditions that ensure the existence of a continuum limit for the posteriors $\mu_N$, the graph pCN algorithm has a uniform spectral gap in Wasserstein distance. 
 	\end{Result}
 	The result hinges on the continuum limit of posteriors discussed in Section \ref{sec:postcontinuumlimit} and on the use of the graph pCN algorithm, which exploits it. Standard MCMC algorithms based on random walks or Langevin dynamics fail to satisfy a uniform spectral gap. The proof is based on a weak Harris theorem \cite{hairer2011asymptotic} that provides necessary conditions for the existence of a Wasserstein spectral gap. An $L^2$ spectral gap can be obtained as a corollary.

 	\section{Posterior Contraction}\label{sec:posteriorcontraction}
 	
 	In Section \ref{sec:postcontinuumlimit} we studied convergence of posteriors $\mu_N$ towards their continuum limit as $N\rightarrow\infty$ with $n$ fixed. The previous section showed that exploiting this continuum limit is essential in order to design sampling algorithms that scale to large number $N$ of features. In this section, we study the performance of graph-based Bayesian SSL when both $N$ and $n$ tend to infinity. The analysis of this double limit discerns if, and how, unlabeled data enhances the learning performance. We provide an affirmative answer for regression and classification, with a quantitative analysis of the scaling of $N$ with $n$ required to achieve optimal learning performance. 
 	
 	Subsection \ref{ssec:background} formalizes the problem setting and our criterion used to quantify the learning performance. We then show in Subsection \ref{ssec:estimation} how the performance of graph-based Bayesian SSL can be analyzed by bringing together the continuum limit of graph-based priors in Subsection \ref{ssec:priorcontinuumlimit} with the theory of Bayesian nonparametrics. 
 	
 	\subsection{Background}\label{ssec:background}
 	To formalize our setting, recall that we are given labeled data $\{(x_i,y_i)\}_{i=1}^n$ sampled independently from the model \eqref{eq:model} and unlabeled data $\{x_i\}_{i=n+1}^{N_n}$ sampled independently from $\law(X)$. Notice that we have introduced a subscript to the total number $N_n$ of features since we are interested in studying its scaling with respect to $n$. We assume that the labels are generated from a fixed \emph{truth} $f_0$ and aim to study the performance of learning $f_0$ with the graph-based Bayesian approach. For our theory, we will view the truth $f_0$ as a function defined on the manifold from where the features are assumed to be sampled.
 	
 	We will use the notion of \emph{posterior contraction rates} \cite{ghosal2000convergence} to quantify the learning performance. This concept, which we will overview in what follows, provides a rigorous footing for the analysis of Bayesian techniques from a frequentist perspective. We will say that the posteriors $\mu_{N_n}$ contract around $f_0$ with rate $\delta_n$ if, for all sufficiently large $M>0$,
 	\begin{align}
 	\mu_{N_n}\Bigl(f\in\mathbb{R}^{N_n}:\|f-f_0\|_{n}\leq M\delta_n \Bigr) \xrightarrow{n\rightarrow\infty} 1 \label{eq:posterior contraction}
 	\end{align}
 	in probability, where 
 	$$\|f-f_0\|_n^2:=\frac1n \sum_{i=1}^n |f(x_i)-f_0(x_i)|^2.$$ Here again we identify a vector in $\mathbb{R}^{N_n}$ with a function over $\{x_i\}_{i=1}^{N_n}$. The convergence \eqref{eq:posterior contraction} implies that, asymptotically, the sequence of posteriors $\mu_{N_n}$ will be nearly supported on a ball of radius $O(\delta_n)$ around $f_0|_{\{x_1,\ldots,x_{N_n}\}}$. Therefore,  $\delta_n$ characterizes the rate at which the posterior ``contracts'' around $f_0,$ and can be intuitively interpreted as the convergence rate of the posterior distribution towards the truth. As a consequence of the convergence \eqref{eq:posterior contraction}, the point estimator 
 	\begin{align*}
 	\widehat{f}_n:= \underset{g\in\mathbb{R}^{N_n}}{\operatorname{arg\, max }}\, \big[\mu_{N_n}(f\in\mathbb{R}^{N_n}: \|f-g\|_n\leq M\delta_n)\big]
 	\end{align*}
 	converges (in probability) to $f_0$ with the same rate $\delta_n$. This observation provides convergence rates for Bayesian point estimators that can be compared with the optimal rates from the minimax theory of statistical inference. 
 	
 	\subsection{Performance and Unlabeled Data}\label{ssec:estimation}
 	We have the following result, formalized in \cite[Theorem 2.1]{sanz2020unlabeled}.
 	\begin{Result}\label{result:posterior contraction}
 		Under a manifold assumption, optimal posterior contraction rates can be achieved if $N_n \gtrsim n^{2m}$. 
 	\end{Result}
 	The result suggests that unlabeled data help and gives a quantitative required scaling of unlabeled and labeled data. We will now illustrate the main ideas behind it. First of all, in order for the unlabeled data to help, there should be some correlation between the truth $f_0$ and the marginal distribution of $X$. This then relates back to the manifold assumption that we have used throughout. The intuition is then that if the truth $f_0$ is a smooth function over the manifold, a better understanding of the underlying geometry through the unlabeled data may improve the learning of $f_0$. This, in terms of our graph-based prior, is reflected by the fact that it is constructed using all of the features. Since the graph-based prior approximates an underlying continuum prior on the manifold, incorporating the unlabeled data allows one to get a better approximation at the level of the prior, which leads to better learning performance at the level of the posterior.

  Another important ingredient in our analysis is that the continuum prior gives optimal learning performance.  Recall that the continuum prior is the Mat\'ern type GP as in \eqref{eq:MGP manifold} or \eqref{eq:KL continuum prior} and $s$ characterizes the smoothness properties of the sample paths. It turns out that if the truth $f_0$ is $\beta$-regular (belonging to a Besov-type space $B_{\infty,\infty}^{\beta}$), then the posteriors with respect to the continuum prior with parameter $s=\beta+m/2$ contract around $f_0$ with rate $n^{-\beta/(2\beta+m)}$ (up to logarithmic factors), which is the minimax optimal rate of estimating a $\beta$-regular function. The key is that, for optimal performance, the prior smoothness parameter $s$ needs to match (up to an additive constant that depends only on the intrinsic dimension) the smoothness $\beta$ of the truth $f_0.$ This agreement is also needed when working with Mat\'ern GPs on Euclidean spaces.
 	
 	Now the final step is to combine the above two main observations:
 	\begin{enumerate}
 		\item Graph-based prior approximates the continuum prior.
 		\item The continuum prior gives optimal posterior contraction rates.
 	\end{enumerate}
 	A result from the Bayesian nonparametrics literature then implies that if the graph-based prior approximates the continuum prior sufficiently well (satisfying an error rate of $n^{-1}$  in an $L^{\infty}$ version of the $d_{TL^2}$ metric introduced in Section \ref{sec:postcontinuumlimit}), then the graph-based prior gives the same posterior contraction rates as the continuum prior, which is again optimal. Therefore the remaining piece is to quantify the approximation error of the continuum prior by the graph-based prior, which is shown to be on the order of $N_n^{-1/2m}$. Therefore the scaling in Result \ref{result:posterior contraction} is obtained by matching $n^{-1}$ and $N_n^{-1/2m}$. The message is that the convergence rate of the graph-based prior suffers from the curse of dimensionality, which is not surprising since the resolution of the $x_i$'s scales like $N_n^{-1/m}$. But the abundance of the unlabeled data alleviates such issue and leads to an accurate approximation of the underlying continuum prior, based on which optimal performance can be achieved.

 	\section{Summary and Open Directions}\label{sec:open}
 	In this article we have overviewed the graph-based Bayesian approach to SSL. We have emphasized how the study of continuum limits provides a rigorous foundation for the design of prior distributions and sampling algorithms with large number of features, and is also a key ingredient in the statistical analysis of posterior contraction. The  foundations of graph-based Bayesian learning are still emerging, and we expect that future contributions will require the development and the  synergistic use of a broad range of mathematical tools, including topology, calculus of variations, spectral graph theory, ergodicity of Markov chains, optimal transport, numerical analysis, and Riemannian geometry. We conclude this article with some theoretical, methodological, and applied open directions. 
 	
 	\subsection{Theory}
 	The uniform spectral gap of the graph pCN algorithm ensures its independent rate of convergence in the limit $N\to \infty$ with fixed number $n$ of labels. However, the rate of convergence of this algorithm would deteriorate in the joint limit $N, n \to \infty.$ The exploration of MCMC algorithms that scale in this joint limit is an interesting open direction. The contraction of the posterior distribution in this regime has been discussed in Section \ref{sec:posteriorcontraction}. Existing results assume an \emph{a priori} known smoothness of the labeling function in order to achieve optimal contraction rates. We believe these results can be extended to achieve \emph{statistical adaptivity}: the smoothness could, in principle, be inferred without hindering the contraction rate. This is an interesting theoretical question, which may also lead to the design of more flexible prior models. Finally, the manifold assumption that our continuum limits rely on is an idealization of the intuitive idea that feautures often contain some low dimensional structure while living in a high dimensional ambient space. In applications, however, data are noisy and it is important to ensure that algorithms designed under a manifold assumption are not sensitive to small perturbations in the data.  In this regard, the paper \cite{ruiyilocalregularization} explores how performing local averages of noisy features can improve the learning performance on noisy point clouds. In addition to relaxing the manifold assumption to account for noisy data, it would be interesting to further develop mathematical foundations for graph-based Bayesian SSL under a cluster assumption (which says that data belonging to the same cluster tend to share the same label).

 	\subsection{Methodology}
 	The design of graph-based prior GPs in SSL takes inspiration from, and shares ideas with, the design of GPs in spatial statistics, where numerous techniques have been developed to enhance the scalability of GP methodology to large datasets. Some of these connections are investigated in \cite{sanz2020spde}, but there are still numerous opportunities for cross-pollination of ideas. For instance, \cite{sanz2021finite} analyzes the finite element approach from spatial statistics using the techniques outlined in Section \ref{sec:posteriorcontraction}. A related topic that deserves further research is the modeling of flexible nonstationary graph-based GPs by appropriate choice of graph-Laplacian and similarities between features. Finally, an important asset of the Bayesian perspective is its ability to provide uncertainty quantification. However, how best to utilize the Bayesian probabilistic framework in the SSL context also requires further research. We envision new opportunities to develop active learning strategies for the adaptive labeling of features. 
 	
 	\subsection{Applications}
 	The ideas and techniques that underpin the foundations and algorithms outlined in this article are bound to be useful beyond the SSL regression and classification problems that have been our focus. Graph-based Bayesian techniques can find application, for instance,  in nonlinear inverse problems. In this direction, \cite{harlim2021graph,harlim2020kernel} investigate PDE-constrained inverse problems on manifolds, where both the prior distribution and the likelihood function involve differential operators supplemented with appropriate boundary conditions. Graphical approximations of these operators call for new continuum limit analyses.

 	\section*{Acknowledgment} Research of the authors described in this review was funded by NSF and NGA through the grants DMS-2027056 and  DMS-2005797. The work of DSA was also partially funded by the NSF grant DMS-1912818, and by FBBVA through a start-up grant.

 	\bibliography{ExampleRefs}
 	
 \end{document}